\documentclass[11pt]{article}

\usepackage[preprint]{acl}

\usepackage{times}
\usepackage{latexsym}
\usepackage{booktabs}
\usepackage{amsmath}

\usepackage[T1]{fontenc}

\usepackage[utf8]{inputenc}

\usepackage{microtype}

\usepackage{inconsolata}

\usepackage{graphicx}
\usepackage{multirow}
\title{Branch-and-Browse: Efficient and Controllable Web Exploration \\ with Tree-Structured Reasoning and Action Memory}

\author{
 \textbf{Shiqi He\textsuperscript{1}},
 \textbf{Yue Cui\textsuperscript{2}},
 \textbf{Xinyu Ma\textsuperscript{3}},
 \textbf{Yaliang Li\textsuperscript{2}},
 \textbf{Bolin Ding\textsuperscript{2}},
 \textbf{Mosharaf Chowdhury\textsuperscript{1}}
\\
 \textsuperscript{1}University of Michigan,
 \textsuperscript{2}Alibaba Group,
 \textsuperscript{3}McMaster University
\\
\texttt{\{shiqihe, mosharaf\}@umich.edu} 
\\
\texttt{\{ciwei.cy, yaliang.li, bolin.ding\}@alibaba-inc.com} 
\\
\texttt{\{ma209\}@mcmaster.ca} 
}

\begin{document}
\maketitle
\begin{abstract}

Autonomous web agents powered by large language models (LLMs) show strong potential for performing goal-oriented tasks such as information retrieval, report generation, and online transactions. These agents mark a key step toward practical embodied reasoning in open web environments. However, existing approaches remain limited in reasoning depth and efficiency: vanilla linear methods fail at multi-step reasoning and lack effective backtracking, while other search strategies are coarse-grained and computationally costly. 
We introduce \emph{Branch-and-Browse}, a fine-grained web agent framework that unifies structured reasoning-acting, contextual memory, and efficient execution. It (i) employs explicit subtask management with tree-structured exploration for controllable multi-branch reasoning, (ii) bootstraps exploration through efficient web state replay with background reasoning, and (iii) leverages a page action memory to share explored actions within and across sessions. 
On the WebArena benchmark, Branch-and-Browse achieves a task success rate of 35.8\% and reduces execution time by up to 40.4\% relative to state-of-the-art methods. These results demonstrate that Branch-and-Browse is a reliable and efficient framework for LLM-based web agents. Code is available at \url{https://github.com/SymbioticLab/Branch-and-Browse}.

\end{abstract}
\section{Introduction}
\label{sec:introduction}


Recent advances in large language models (LLMs) have enabled the development of intelligent agents capable of perceiving webpages, reasoning over content, and acting to automate end-to-end tasks such as online shopping, content management, and issue tracking. LLM-based agents have shown great potential in automating repetitive and programmatic workflows, thereby alleviating human effort in real-world web interaction scenarios \citep{gao2024large, wang2024survey, xi2025rise, yang2024if}. These capabilities stem from LLMs’ strong generalization in perception, reasoning, and planning, acquired through large-scale pre-training and instruction tuning \cite{achiam2023gpt, dubey2024llama, hurst2024gpt, yang2025qwen3}.

\begin{figure}[t]
  \includegraphics[width=\columnwidth]{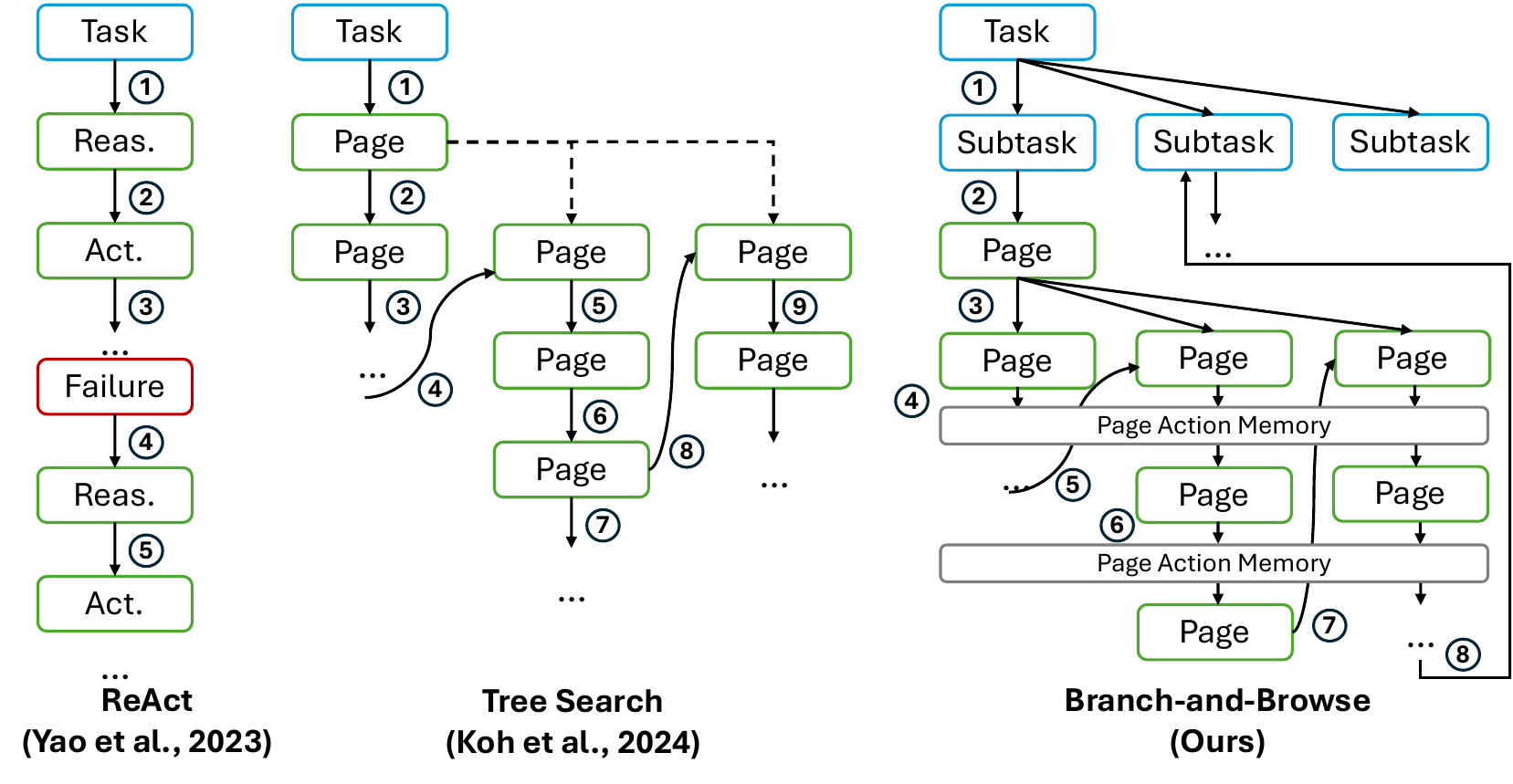}
  \caption{Comparison of web agent strategies. The left shows linear prompting (ReAct) with no backtracking, the middle shows inefficient tree search exploration, and the right illustrates our Branch-and-Browse framework, which enables fine-grained, memory-guided, and efficient multi-branch reasoning.}
  \label{fig:intro}
  \vspace{-10px}
\end{figure}

Unlike static natural language tasks such as summarization or dialogue \cite{zhang2025systematic,chen2025towards,cui2025efficient}, web environments pose greater challenges due to their diverse, context-dependent action choices/function calling (e.g., \texttt{click}, \texttt{type}, \texttt{scroll}) \cite{masterman2024landscape,qu2025tool,cui2025enhancing} and partial observability \cite{ning2025survey,gao2025agentscope}. Solving such tasks requires long-horizon reasoning, explicit exploration, and efficient backtracking across dynamic interfaces. Yet LLM-based agents often fail to align their learned knowledge with task-specific observations and actions, leading to reasoning drift and execution errors. As a result, on benchmarks such as WebArena and VisualWebArena \citep{zhou2023webarena, koh2024visualwebarena}, current web agents significantly underperform human users, revealing fundamental challenges in integrating reasoning and decision-making within interactive, real-world web environments. 

To cope with these challenges, recent research has explored various strategies that combine reasoning and action generation.
Linear prompting-based approaches (e.g., \emph{ReAct}, as illustrated in Figure~\ref{fig:intro}) \citep{yao2023react} follow a single reasoning–action trajectory, which limits their ability to recover from early mistakes. Once an incorrect click or form submission is made—such as navigating to the wrong product page or misfilling a query field—the agent cannot efficiently backtrack, often restarting the entire sequence. In contrast, search-based methods such as \emph{Tree Search} \citep{koh2024tree} expand multiple states, thereby improving task completion rates through exploration of multiple pathways. Despite this advantage, these methods suffer from coarse granularity and high computational costs, as they explore each branch independently without effectively utilizing shared contextual information—such as pages previously visited or patterns of failed interactions. These limitations result in redundant exploration and inefficient decision-making processes.


To tackle these limitations, we propose \emph{Branch-and-Browse}, a fine-grained exploration framework that integrates structured reasoning-acting, contextual memory, and efficient execution. Our method provides: (i) \emph{fine-grained structured reasoning} through a subtask manager and tree-structured exploration that enable controllable multi-branch reasoning and principled backtracking; (ii) \emph{web state replay}, which efficiently recover the next branch for exploration, and \emph{background reasoning}, which leverages offline evaluations of unexplored nodes to prune unpromising branches, prioritize actionable steps, and significantly improve the effective branching factor; (iii) a \emph{page action memory} that records explored actions and outcomes, sharing them across branches to reduce redundancy and accelerate decision-making.

We summarize our contributions as follows:

\begin{itemize}
\item We propose \emph{Branch-and-Browse}, a novel subtask-aware, tree-structured exploration framework that enables fine-grained, structured reasoning-acting, supporting controllable multi-branch exploration, principled backtracking, and background branch evaluation to enhance reasoning flexibility and exploration efficiency.
\item We design a \emph{page action memory} mechanism that addresses contextual fragmentation by maintaining page-level summaries and cross-branch exploration histories.  \emph{Background reasoning} mechanism is also introduced to tackle the challenge of exploration efficiency in dynamic branching tasks. 
\item Our approach attains state-of-the-art performance on WebArena, with a 35.8\% task success rate and up to a 40.4\% reduction in execution time relative to existing methods, demonstrating its ability to balance reasoning depth and multi-branch exploration efficiency in dynamic web environments. 
\end{itemize}

\begin{table}[t]
\footnotesize
\centering
\caption{Action space $\mathcal{A}$ for web interaction. Each action $a \in \mathcal{A}$ defines a permissible atomic operation on a web page or browser context. This represents a subset of functions provided by the Playwright MCP toolkit.}
\label{tab:action_space}
\begin{tabular}{ll}
\toprule
\textbf{Action} & \textbf{Description} \\
\midrule
\texttt{NAVIGATE(url)} & Navigate to a URL. \\
\texttt{NAVIGATE\_BACK()} & Move backward. \\
\texttt{NAVIGATE\_FORWARD()} & Move forward. \\
\texttt{CLICK(element)} & Click an element. \\
\texttt{TYPE(element, text)} & Type text into a field. \\
\texttt{SELECT(element, option)} & Select an option. \\
\texttt{HOVER(element)} & Hover over an element. \\
\texttt{DRAG(source, target)} & Drag an element. \\
\texttt{PRESS\_KEY(key)} & Press a keyboard key. \\
\texttt{TAB\_NEW()} & Open a new tab. \\
\texttt{TAB\_SELECT(id)} & Switch to a specific tab. \\
\texttt{TAB\_CLOSE(id)} & Close a specific tab. \\
\bottomrule
\end{tabular}
\end{table}

\begin{figure*}[t]
  \centering
  \includegraphics[width=1.98\columnwidth]{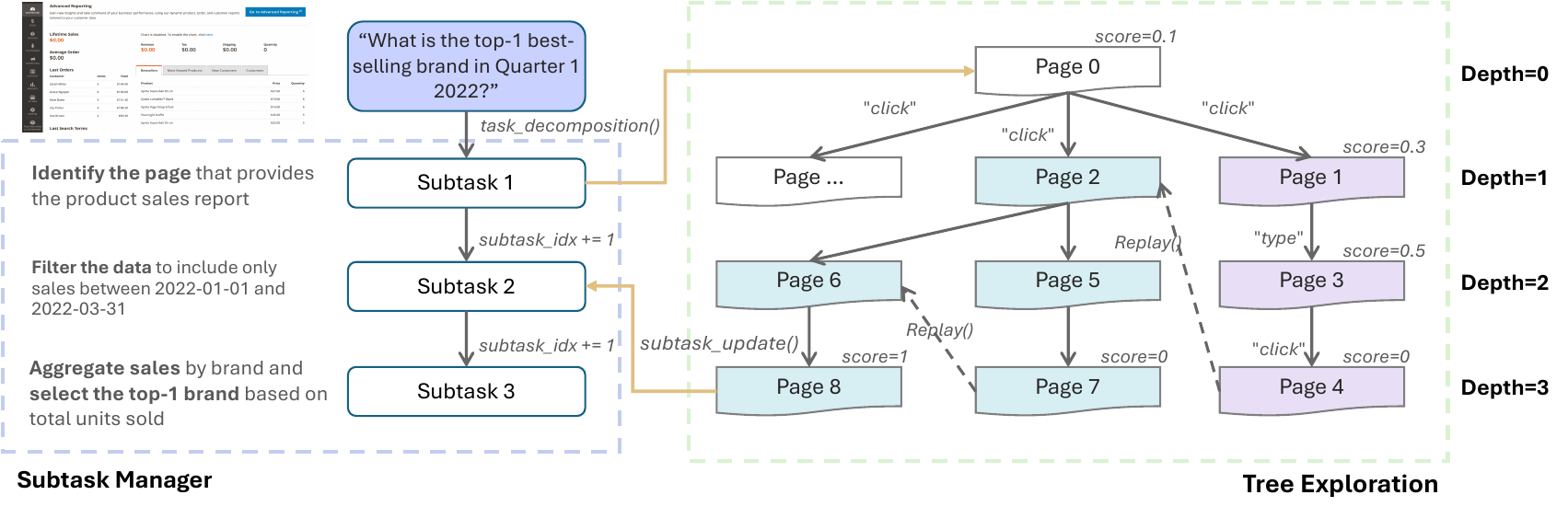}
  \caption{\textbf{Pipeline of the Branch-and-Browse framework.} 
Given an example task \citep{zhou2023webarena}, the \emph{subtask manager} first decomposes the goal into three subtasks: (1) identify the page providing the product sales report, (2) filter data for the Q1 2022 period, and (3) aggregate brand sales to select the top-1 brand. Each subtask is explored through a tree-structured exploration consisting of iterative \texttt{Reason–Act–Evaluation} cycles. Nodes represent visited pages, and edges denote executed system actions (e.g., \texttt{click}, \texttt{type}). Depth corresponds to exploration steps, while replay links (dashed) enable efficient backtracking and reuse of prior context. As shown, successful branches (e.g., Page~8) trigger \texttt{subtask\_update()} to advance to the next subgoal, allowing the agent to progress hierarchically through subtasks while avoiding redundant exploration.
}
  \label{fig:overview}
  \vspace{-10px}
\end{figure*}

\section{Background}
\label{sec:related}
\textbf{Web Agent Environment.} Web agents are designed to perform goal-directed tasks through programmatic interaction with real or simulated web interfaces \cite{ning2025survey}. Recent implementations leverage large language models (LLMs) as the core reasoning component \citep{gur2023real, boisvert2024workarena++, drouin2024workarena, zheng2024gpt}, but these agents form part of a broader system that interfaces with websites through browser automation. Following \citet{zhou2023webarena}, we formalize web agents within a sequential decision-making framework, where a task is specified as a natural language intent $i$ (e.g., “find all shoes under 50 dollars and add them to my wishlist”). At each time step $t$, the agent observes $o_t \in \mathcal{O}$ and issues an action $a_t \in \mathcal{A}$ according to a policy $\pi(a_t | i, o_t, a_{1:t-1}, o_{1:t-1})$. Executing $a_t$ transitions the environment to a new state $s_{t+1} = Trans(s_t, a_t)$ with an updated observation $o_{t+1}$. The action space $\mathcal{A}$ includes operations analogous to human interactions—element selection (\texttt{click}), text input (\texttt{type}), URL navigation (\texttt{goto}), and tab management—implemented via automation frameworks such as Playwright.\footnote{\url{https://playwright.dev}} The effectiveness of an agent can be assessed through a reward function $r(a_{1:T}, s_{1:T})$ that programmatically verifies whether the final state satisfies the task intent. Table~\ref{tab:action_space} summarizes the action primitives used in our environment.


\textbf{Reasoning and Action Generation.} Recent studies have explored diverse strategies for integrating reasoning and action generation in web agents. Early approaches such as \emph{ReAct} \citep{yao2023react} follow a single reasoning–action trajectory, lacking robust backtracking, while \emph{Tree Search} \citep{koh2024tree} improves exploration through branching but remains coarse-grained and computationally costly. Subsequent frameworks like \emph{ScreenAgent} \citep{niu2024screenagent}, \emph{OS-Genesis} \citep{sun2024genesis}, \emph{WebDreamer} \citep{gu2024your}, and \emph{AgentScope} \citep{gao2025agentscope} introduce structured subtasks, reflection, and predictive reasoning to enhance planning. However, no unified framework currently integrates these strengths—structured exploration, subtask awareness, and contextual reasoning—into a fine-grained architecture capable of maintaining control while mitigating exploration inefficiency.

\textbf{Challenges and Opportunities.} Despite recent progress, real-world web tasks remain dynamic and partially observable, requiring agents to reason over long horizons and adapt their strategies based on intermediate feedback. Three main challenges arise: (i) \textit{branching uncertainty}, where multiple paths diverge from the goal—handled by Branch-and-Browse through a subtask manager and tree-structured planning for controlled exploration and backtracking; (ii) \textit{exploration inefficiency}, caused by redundant branch expansion—alleviated via efficient web-state replay and background reasoning; and (iii) \textit{contextual fragmentation}, where past interaction histories are lost—addressed by a page action memory that summarizes explored actions and shares context across branches.


\vspace{-2px}

\section{Design}
\label{sec:method}

We formulate long-horizon web interaction as a structured search problem over a dynamic environment. Given a natural language intent $i$ and the corresponding initial observation $o_0$, the agent explores possible trajectories under a sequential decision process guided by structured planning, subtask-aware control, and page-level memory. A trajectory $\tau = (o_0, a_0, o_1, a_1, \dots, o_T)$ is generated by alternately selecting actions $a_t \in \mathcal{A}$ according to the policy $\pi(a_t | i, o_t, a_{1:t-1}, o_{1:t-1})$ and applying the environment transition operator $T$ such that $o_{t+1} = Trans(o_t, a_t)$. The objective is to reach a terminal observation $o_T$ that satisfies the task goal $\mathcal{G}(i)$, indicating successful completion of the user intent.

\subsection{Fine-Grained Structured Planning}
Branch-and-Browse performs structured planning through hierarchical task decomposition and tree-based exploration.
Given a global instruction $i$, the agent first generates subtasks $\{u_1, \dots, u_K\}$ via \texttt{task\_decomposition()}, then explores each through iterative \texttt{Reason–Act–Evaluation} cycles.
During exploration, active subtasks are adaptively refined using \texttt{subtask\_update()} based on the current page context and progress (Figure~\ref{fig:overview}).

\paragraph{Subtask Manager.} Similar to general agentic tasks \cite{plaat2025agentic,acharya2025agentic,xu2025comprehensive}, complex web instructions often require sequential reasoning over intermediate objectives. 
The \emph{subtask manager} provides structured workflow control by maintaining an active subtask $u_k$ at each iteration. 
Each subtask is defined by an objective function $\mathcal{O}(u_k)$ and a success predicate $\mathcal{A}(u_k)$ that indicates whether progress toward the goal has been achieved. 
Initially, the set of subtasks $\{u_1, \dots, u_K\}$ is generated by
\[
\{u_1, \dots, u_K\} = \texttt{task\_decomposition}(i),
\]
where $i$ is the natural language task intent. 
Since this decomposition occurs before any real page context is available, some subtasks may not align with the actual site structure—e.g., a subtask expecting a “sales report” section on an e-commerce homepage where such content does not exist. 
To address this, after each round of exploration, the manager performs a contextual update using
\[
u_k \leftarrow \texttt{subtask\_update}(u_k, o_t, \tau_{\leq t}),
\]
even when no progress is made. 
This adaptive refinement allows the agent to reinterpret or reformulate the current subtask based on observed evidence—such as replacing an unreachable objective with a semantically related alternative discovered through browsing. 
If $\mathcal{A}(u_k)$ is satisfied, the subtask is marked as complete and the next subtask $u_{k+1}$ is activated. 
This dynamic adjustment prevents dead-end reasoning and ensures that exploration remains grounded in the real page context.

\paragraph{Tree Exploration.}

The exploration process is formulated as a search over a tree $\mathcal{T}$, where each node represents a visited web page (state) and each edge corresponds to a candidate system action $a_t \in \mathcal{A}$. 
The agent maintains a frontier $\mathcal{F}$ of expandable nodes, each associated with a value estimate guiding prioritization~\cite{koh2024tree}. 
At each iteration, the agent selects the node with the highest utility value $v$ from the frontier:
\[
(o, v) \leftarrow \arg\max_{(o', v') \in \mathcal{F}} v'.
\]
Each $v$ is given by an LLM-as-a-judge evaluator following the tree-search scoring principle~\cite{koh2024tree}, but conditioned on the \emph{active} subtask $u_k$, the current observation and browsing history, and page action memory so the judge can distinguish redundant retries from new progress. The structured model outputs are mapped to a scalar $v \in [0,1]$. For each subtask we store the best $v$ seen on any explored prefix and update it when a descendant improves that level, so frontier ordering tracks subtask-local progress rather than a single global score alone.
The selected node $o$ is then expanded by generating a set of candidate actions $\{a_i\}_{i=1}^{b}$, where $b$ is the branching factor controlling exploration breadth. 
Each action yields a successor observation $o^{(i)} = Trans(o, a_i)$, which inherits contextual memory and the current subtask state from its parent.


Exploration proceeds through iterative \texttt{Reason–Act–Evaluation} cycles, during which the agent reasons about the next action, executes it, and evaluates the resulting page state. 
Branches that achieve subtask goals trigger \texttt{subtask\_update()} to transition to the next subgoal, while low-value or repetitive branches are pruned from the frontier. 
Replay links (dashed edges in Figure~\ref{fig:overview}) allow fast backtracking and reuse of previously visited contexts, preventing redundant page loads. 
The process continues until a node satisfies the overall task goal $\mathcal{G}(i)$ or the exploration budget $c$ is consumed. 
This hierarchical, subtask-aware search structure ensures controllable exploration depth, efficient reuse of prior context, and robust completion of multi-step web tasks.



\begin{figure*}[t]
  \centering
  \includegraphics[width=1.98\columnwidth]{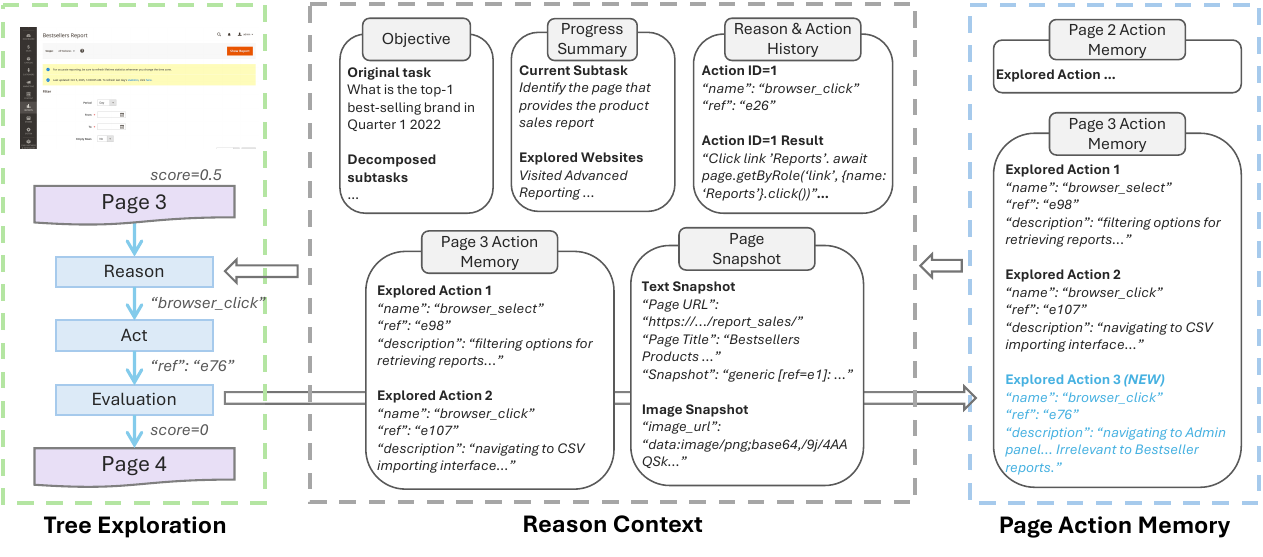}
  \caption{\textbf{Example of reason context in branch-and-browse.} During tree exploration, the agent on Page~3 reasons to execute a web interaction (\texttt{browser\_click}), acts on the referenced element, and evaluates the outcome (score~=~0), leading to Page~4. The page-level context records the task objective, progress summary, reason–action history, and page snapshot. The \texttt{Page~3 Action Memory} is updated by appending the new action (highlighted in blue) while marking it as irrelevant to the current goal. This cached record guides future reasoning, allowing the agent to avoid redundant exploration and efficiently backtrack during replay.
}
  \label{fig:context}
  \vspace{-10px}
\end{figure*}

\subsection{Exploration Acceleration}
To enhance the efficiency of structured planning, Branch-and-Browse incorporates two complementary acceleration mechanisms: \textit{nearest-URL state replay} and \textit{background reasoning}.  
These components reduce redundant execution, enable fast recovery from failed trajectories, and improve search efficiency by prioritizing promising branches without exhaustive expansion.

\paragraph{Nearest-URL State Replay.}  
Backtracking is frequently required when exploration paths fail or deviate from the goal.  
A naive approach of replaying only by navigating directly to a saved URL often misses crucial intermediate actions—such as form filling or tab switching—that are necessary to reconstruct the precise state.  
Conversely, fully re-executing the entire trajectory $\tau = (o_0, a_0, o_1, \dots, o_t)$ can be prohibitively time-consuming, especially for long sequences involving multiple page loads.  
To balance these trade-offs, Branch-and-Browse employs a nearest-URL replay strategy: to revisit an earlier state $o_j$ ($j < t$), the agent restores the closest cached URL $url_c$ ($c \leq j$) and replays only the remaining actions $(a_c, \dots, a_{j-1})$:
\[
\textsc{Replay}(\tau, j) = \textsc{Load}(url_c) \xrightarrow{a_c} \dots \xrightarrow{a_{j-1}} o_j.
\]
This hybrid approach preserves essential interaction context while avoiding redundant re-execution, achieving both accuracy and efficiency in backtracking.

\paragraph{Background Reasoning.}  
To further reduce redundant interactions, unexplored frontier nodes $o \in \mathcal{F}$ are evaluated through offline reasoning without active execution.  
Each node is represented by a context $C(o)$ derived from its DOM snapshot $o$, and URL.  
The reasoning model analyzes $C(o)$ to infer plausible next actions and their relevance to the current subtask.  
When the inferred action corresponds to a \texttt{click} operation explicitly linked to a valid URL in the page text, the node is pre-expanded in the background by simulating this transition, i.e., following the referenced URL to create a new state.  
For all other actions (e.g., \texttt{type}, \texttt{select}), execution is deferred until the node becomes actively focused in the main exploration branch, since such interactions require a live web context.  
This selective pre-expansion allows efficient evaluation of deterministic navigation steps, thereby accelerating overall search without compromising correctness.

\subsection{Page Action Memory}

Global context that logging all agent steps across branches often becomes redundant and entangled, making retrieval inefficient and replay unreliable \cite{yao2023react,koh2024tree}.  
To address this, Branch-and-Browse introduces a \emph{page action memory} module that maintains structured reasoning and interaction records at the granularity of each visited page URL.  
This design allows the agent to retrieve, summarize, and update information efficiently during both online reasoning and replay.  
As illustrated in Figure~\ref{fig:context}, each reasoning context consists of the following components:

\begin{itemize}
    \item \textbf{Objective:} Stores both the global task intent and the currently active subtask.  
    This maintains hierarchical reasoning context, enabling consistent goal alignment throughout exploration.

    \item \textbf{Progress Summary:} Provides a concise textual overview of exploration progress—listing visited sites, their relevance to the subtask, and key intermediate findings.  
    These summaries guide subsequent action generation and serve as inputs for background reasoning.

    \item \textbf{Reason–Action History:} Records the sequence of reasoning and executed system actions (e.g., \texttt{click}, \texttt{select}, \texttt{type}), together with element references and post-action results.  
    Each record maintains \texttt{\{ActionID, name, ref, result\}}, supporting deterministic replay and efficient backtracking during nearest-URL restoration.

    \item \textbf{Page Snapshot:} Captures both textual and visual representations of the page.  
    The text snapshot includes the URL, title, and compressed DOM structure for contextual grounding, while the image snapshot stores a base64-encoded screenshot reference for multimodal reasoning and verification.

    \item \textbf{Action Memory:} Maintains a structured log of all actions attempted on the page, together with metadata and success indicators.  
    Each new action (e.g., \texttt{click} → “Admin panel”) is appended while marking redundant or irrelevant ones as low priority.  
    This evolving cache prevents repeated exploration of unproductive branches and serves as a shared knowledge source for background reasoning and replay.
\end{itemize}

After each \texttt{Reason–Act–Evaluation} cycle, the memory for the current page is updated and serialized into an external cache.  
When the agent revisits the same URL, the stored summary and cached actions are reloaded to reconstruct local context, enabling consistent reasoning, efficient replay, and branch pruning.  
Moreover, during subsequent task decomposition, the visited actions recorded in the page memory are incorporated into the reasoning context, allowing the agent to refine subtasks based on explored website structures and improve decomposition accuracy.  
By structuring memory at the page level and coupling it with system-level command traces, the agent can (i) avoid redundant exploration, (ii) compress historical traces into meaningful summaries, and (iii) recover relevant state rapidly during nearest-URL replay or background reasoning.


\begin{table*}[!t]
    \centering
    \caption{Comparison of the success rate (SR) of our method with baseline agents on the WebArena (WA) benchmark. Published baselines are shown in the upper block, while tree-search-based methods, our method, and a SteP-augmented hybrid are reported in the lower block.}
    \label{tab:main_results}
    \resizebox{\textwidth}{!}{
    \begin{tabular}{lcccccccc}
        \toprule
        \textbf{Method} & \textbf{Model} & \textbf{SR (\%)} & \textbf{Shopping}  & \textbf{Shopping Admin}  & \textbf{GitLab}  & \textbf{Map}  & \textbf{Reddit}  & \textbf{Multisite} \\
        \midrule
        WebArena~\cite{zhou2023webarena} & GPT-4-Turbo & 16.5 & 16.6 & 15.9 & 10.0 & 22.9 & 21.7 & 16.7 \\
        BrowserGym~\cite{drouin2024workarena} & GPT-4o & 23.5 & -- & -- & -- & -- & -- & --\\
        \midrule
        AutoWebGLM\cite{lai2024autowebglm} & ChatGLM3-6B & 18.2 & -- & -- & -- & -- & -- & --\\
        AutoEval~\cite{pan2024autonomous} & GPT-4 & 26.9 & 39.6 & 20.9 & 25.0 & 27.5 & 20.8 & 16.7\\
        SteP~\cite{sodhi2023step} & GPT-4-Turbo & 33.3 & 33.2 & 32.4 & 26.7 & 35.8 & 52.8 & 12.5 \\  
        AWM~\cite{wang2024agent} & GPT-4 & 35.5 & 32.1 & 29.1 & 35.0 & 42.2 & 54.7 & 18.8 \\
        API Hybrid Agent~\cite{song2024beyond} & GPT-4o & 38.9 & 25.7 & 41.2 & 44.4 & 45.9 & 51.9 & 16.7 \\        
        WebPilot~\cite{zhang2025webpilot} & GPT-4o & 37.2 & 36.9 & 24.7 & 39.4 & 33.9 & 65.1 & * \\    
        AgentOccam-Judge~\cite{yang2024agentoccam} & GPT-4-Turbo & 45.7 & 43.3 & 46.2 & 38.9 & 52.3 & 67.0 & 16.7 \\
        \midrule
        Tree Search~\cite{koh2024tree} & GPT-4o & 19.2 & 27.8 & 16.5 & 13.9 & 26.6 & 11.3 & 16.7\\
        Branch-and-Browse (Ours) & GPT-4o & 35.8& 34.6&26.4 &36.7&46.8&50.9 &18.8 \\
        Branch-and-Browse + SteP (Ours) & GPT-4o & 39.8 & 41.2 & 32.4 & 37.8 & 47.7 & 54.7 & 18.8 \\
        \bottomrule
        
    \end{tabular}}
    \vspace{-5px}
\end{table*}

\begin{figure*}[h]
    \centering
    \includegraphics[width=0.98\linewidth]{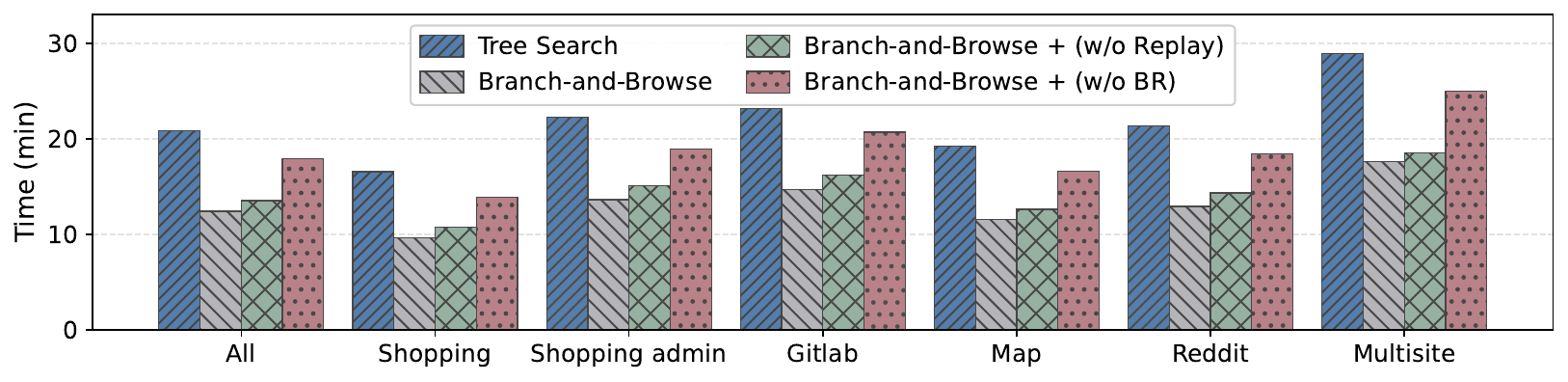}
    \caption{Ablation results on WebArena (812 tasks): average time per successful task across all sites. Failed tasks are excluded since they often fall into repetitive exploration loops, leading to disproportionately high and uninformative time consumption.}
    \label{fig:ablation}
    \vspace{-15px}
\end{figure*}

\section{Evaluation}
\label{sec:evaluation}
In this section, we systematically evaluate our approach in a realistic environment for testing autonomous language agents on complex web tasks. Our evaluation is designed to answer the following questions:
\begin{itemize}
    \item \textbf{Q1:} How does Branch-and-Browse perform compared to existing web agent baselines across different website categories and task types?
    \item \textbf{Q2:} What are the individual and combined contributions of the proposed acceleration mechanisms—nearest-URL replay and background reasoning—to efficiency?
    \item \textbf{Q3:} How do key hyperparameters, i.e., depth and branching factor, affect the performance and stability of the framework?
\end{itemize}
We first present the experimental setup, followed by the main results, ablations, and sensitivity analyses.

\subsection{Experimental Setup}
\paragraph{Datasets.} We evaluate our method on the WebArena (WA) benchmark \cite{zhou2023webarena}, a realistic web environment designed to assess the performance of autonomous language agents on complex web-based tasks. WebArena bridges the gap between synthetic evaluation settings and real-world scenarios by providing fully functional websites across multiple domains, including an e-commerce website (OneStopShop), GitLab, Reddit, an online store content management system (Shopping Admin), a map, and an English Wikipedia. The benchmark consists of 812 long-horizon tasks instantiated from 241 task templates. We additionally evaluate on VisualWebArena~\cite{koh2024visualwebarena}, a multimodal dataset built on the same WebArena-style stacks, including images paired with instructions.

\paragraph{Baseline Methods.} We divide the baselines into three categories. WebArena \cite{zhou2023webarena} and BrowserGym \cite{drouin2024workarena} are vanilla ReAct agents that operate within a single reasoning–action loop. Policy-based strategies such as SteP \cite{sodhi2023step} and AgentOccam-Judge \cite{yang2024agentoccam} decompose complex web tasks into modular or hierarchical policies for more structured control. In contrast, Tree Search \cite{koh2024tree} and our Branch-and-Browse agent are search-based strategies that explicitly explore and evaluate multiple reasoning trajectories, orthogonal to policy-based methods. All evaluations are using isolated server instances with standardized browser instrumentation. Agents interact via natural-language instructions and web-navigation actions, and performance is measured by success rate (SR)—the percentage of tasks achieving the goal state.

\paragraph{Implementation Details.} Our system is implemented on top of the Playwright MCP framework,\footnote{\url{https://github.com/microsoft/playwright-mcp}} which provides browser-level automation for programmatic web interaction, and is integrated within the \texttt{AgentScope} framework \citep{gao2025agentscope} for unified agent orchestration and reasoning. The backend language model is \texttt{gpt-4o-2023-12-12}, serving as the core reasoning and evaluation module throughout task decomposition, action generation, and node scoring. Unless otherwise stated, the tree search uses a depth factor $d = 5$ and branching factor $b = 5$ with an exploration budget of 10 steps per task. All experiments are conducted under identical browser configurations, with page-level memory and background reasoning enabled by default.

\subsection{Main Results}
\paragraph{Results on WebArena.} Table~\ref{tab:main_results} summarizes the performance of Branch-and-Browse compared to the \textit{Tree Search} baseline across 812 tasks.    
Naive strategies with ReAct exhibit limited success due to their lack of structured exploration and contextual recovery, often failing on long-horizon or dynamically changing pages.  Our method achieves substantial gains overall (SR: 35.8\% vs.\ 19.2\%), with particularly strong improvements in domains requiring dynamic reasoning and interactive operations.  
For example, performance increases by 39.6 points on \emph{Reddit} and 22.8 on \emph{GitLab}.  
These domains demand multi-step exploration and contextual consistency, where fine-grained subtask control and replay mechanisms are most beneficial.  
In contrast, the gain on \emph{Multisite} tasks is smaller (i.e., 2.1 points), as such tasks often span multiple websites and require longer trajectories, making them inherently more challenging.  
While several policy-based agents (e.g., SteP, API Hybrid Agent, AgentOccam-Judge) achieve higher absolute scores by leveraging website-specific heuristics or optimized input representations \citep{sodhi2023step, song2024beyond, yang2024agentoccam}, these methods are orthogonal to—and potentially complementary with—our structured, search-based framework.

\paragraph{Combining with SteP.}
Following prior work that composes web agents with additional policy layers to study how modular controllers interact with environment actions \citep{yang2024agentoccam}, we instantiate a \emph{Branch-and-Browse + SteP} hybrid to demonstrate this complementarity empirically rather than only by analogy. The framework is unchanged except at node expansion: instead of a single ReAct-style policy, the agent invokes SteP~\citep{sodhi2023step}, which selects actions through a small library of prompt-based sub-policies (optionally nested on a stack) but ultimately returns one executable web action per expansion, which is executed and recorded as a tree edge; search, replay, background reasoning, pruning, and page action memory follow the default Branch-and-Browse logic. Table~\ref{tab:main_results} shows that this hybrid raises overall SR from 35.8\% to 39.8\%, with gains on Shopping, Shopping Admin, GitLab, Map, and Reddit and unchanged Multisite performance. The pattern aligns with SteP reducing common interaction mistakes in domains that benefit from structured routines---for instance, map tasks where sub-policies for nearest-place search and directions avoid brittle misuse of generic widgets---while Branch-and-Browse still supplies multi-branch exploration and the acceleration mechanisms absent from policy-only stacks.

\begin{table}[t]
    \centering
    \scriptsize
    \caption{Success rate (SR, \%) on VisualWebArena~\citep{koh2024visualwebarena} by site category and overall.}
    \label{tab:visual_webarena}
    \begin{tabular}{lccccc}
        \toprule
        \textbf{Method} & \textbf{Classifieds} & \textbf{Reddit} & \textbf{Shopping} & \textbf{Overall} \\
        \midrule
        GPT-4o & 18.4 & 17.1 & 20.0 & 18.9 \\
        Tree Search & 26.5 & 20.5 & 29.0 & 26.4 \\
        Branch-and-Browse & \textbf{37.2} & \textbf{25.5} & \textbf{33.8} & \textbf{32.7} \\
        \bottomrule
    \end{tabular}
    \vspace{-8px}
\end{table}

\paragraph{Results on VisualWebArena.}
Beyond text-centric WebArena, we report results on VisualWebArena~\citep{koh2024visualwebarena}, which keeps the same reproducible, self-hosted sites but shifts the observation interface toward multimodal inputs: agents must ground decisions in rendered pages and visual layout, complementing language-only benchmarks. Table~\ref{tab:visual_webarena} compares a GPT-4o reactive baseline, \textit{Tree Search}~\citep{koh2024tree}, and Branch-and-Browse with identical model and budget conventions as our WebArena experiments. Branch-and-Browse improves over \textit{Tree Search} on every reported slice (Classifieds, Reddit, Shopping) and raises overall success rate from 26.4\% to 32.7\%, indicating that subtask-aware tree exploration with replay and page action memory remains effective when observations emphasize screenshots rather than purely textual state abstractions.


\subsection{Ablation Studies}
To analyze the contribution of each efficiency mechanism, we conduct ablation studies on the \textit{Replay} and \textit{Background Reasoning} modules.  
Figure~\ref{fig:ablation} compares the average completion time per task against the \textit{Tree Search} baseline across all domains.  
For a fair comparison, we report the average time only on \emph{successful} tasks, as failed trajectories often enter repetitive loops and inflate runtime without contributing meaningful progress.

Overall, Branch-and-Browse substantially reduces execution time compared to \textit{Tree Search}, lowering the average task time from 20.8 to 12.4 minutes (a 40.4\% reduction).  
When \textit{Replay} is disabled, the average time increases slightly, showing that nearest-URL replay helps avoid redundant re-navigation but contributes less to total efficiency gains.  
In contrast, removing \textit{Background Reasoning} leads to a larger increase in task time, since offline inference overlaps with active exploration and prunes unpromising branches early.  
This difference is expected, as reasoning typically dominates runtime compared to replay operations.  
Together, these results confirm that both mechanisms are effective, with background reasoning providing the greater contribution to overall efficiency.






    

\begin{table}[t]
    \centering
    \footnotesize
    \caption{Sensitivity of success rate and runtime to search depth and branching factor on WebArena. }
    \label{tab:ablation}
    \begin{tabular}{cccc}
        \toprule
        \textbf{Depth $d$} & \textbf{Branch $b$} & \textbf{SR ($\uparrow$)}& \textbf{Time ($\uparrow$)}\\
        \midrule
        0 & 1 & 23.9\% & 8.2 min \\
        \midrule
        \multirow{2}{*}{1}
        & 3 & 25.3\% & 8.7 min\\        
        & 5 & 31.5\% & 10.9 min\\
        \midrule
        \multirow{2}{*}{2}
        & 3 & 30.9\% & 10.7 min\\        
        & 5 & 33.7\% & 11.6 min\\
        \midrule
        3 & 5 & 34.4\% & 11.9 min \\
        \midrule
        5 & 5 & 35.8\% & 12.4 min \\
        \bottomrule
    \end{tabular}
    \vspace{-10px}
\end{table}

\subsection{Sensitivity Studies}
We analyze the sensitivity of Branch-and-Browse to search hyperparameters, including the tree depth ($d$) and branching factor ($b$).  
All experiments are conducted under the same overall search budget as the default setting (equivalent to the number of \texttt{ReAct} steps performed by the baseline), ensuring a fair comparison in total interaction count and model queries.

As shown in Table~\ref{tab:ablation}, when both $d=0$ and $b=1$, the framework degenerates to a linear execution mode identical to \texttt{ReAct} augmented with the page action memory.  
This configuration still benefits from context reuse but lacks acceleration from structured exploration, resulting in a success rate of 23.9\%.  
As we increase either the branching factor or the search depth, performance improves significantly while runtime grows moderately.  
For instance, expanding the search to $b=5$ and $d=2$ yields a success rate of 33.7\% with only a 3.4-minute increase in average task time.  
The gains are largely attributed to the efficiency mechanisms—nearest-URL replay and background reasoning—that enable deeper and broader exploration without linear growth in cost.  
Overall, Branch-and-Browse demonstrates strong robustness: larger search budgets consistently improve success rates, while execution time remains well within practical limits, confirming the scalability and efficiency of the framework.

\section{Related Work}
\textbf{Perception and Web Environmental Understanding}: Modern web agents leverage diverse modalities for environmental perception. Text-based methods \cite{deng2023mind2web,koh2024tree,gur2023real,ma2023laser} utilize HTML metadata and accessibility trees, such as the approach in \cite{koh2024tree}, which takes accessibility trees of web pages as input. Screenshot-based techniques incorporate vision-language models to interpret graphical user interfaces (GUIs) \cite{singh2025trishul,cheng2024seeclick,wu2024atlas,sun2024genesis,zhang2023you,kil2024dual}. For example, SeeClick \cite{cheng2024seeclick} exclusively uses screenshots as observations and enhances grounding through targeted pre-training. Multi-modal approaches \cite{song2024mmac,he2024webvoyager} integrate the complementary capabilities of various modalities. Notable examples include MMAC-Copilot \cite{song2024mmac}, which integrates GPT-4V \cite{openai2023gpt4v} for visual content, and Gemini Vision \cite{li2024mini}, which processes video data—significantly improving multi-modal data handling.

\textbf{Memory and Knowledge Integration.} Advanced web agents utilize both short-term and long-term memory systems to enhance decision-making and task performance. For instance, AutoWebGLM \cite{lai2024autowebglm} models web browsing as a sequential decision-making process, where actions are determined based on the current state and an integrated history of previous webpage states and actions. Similarly, Agent S \cite{agashe2024agent} combines external online web searches for supplementary knowledge with an internal narrative memory to draw on task-specific experiences. This includes generating sub-tasks derived from summaries of both successful and unsuccessful trajectories.

\section{Conclusion}
\label{sec:conclusion}
We introduced Branch-and-Browse, a fine-grained framework for efficient and reliable LLM-based web agents.  
By combining subtask-aware tree exploration, page action memory, and acceleration mechanisms, it enables structured and efficient reasoning.  
On the WebArena benchmark, Branch-and-Browse achieves a 35.8\% success rate and reduces execution time by 40.4\%, demonstrating a scalable path toward practical, high-performing web agents.

\section*{Limitations}
While Branch-and-Browse improves the efficiency and reliability of web agents through structured exploration, it still operates within a single-browser session setting and does not parallelize exploration across multiple branches.  
This limits scalability in highly complex or time-sensitive tasks where concurrent reasoning could further accelerate progress.  
Additionally, our framework primarily focuses on search-based control and has not yet been fully integrated with policy-based approaches, which may offer complementary benefits.  
Future work will explore multi-session and hybrid policy–search integration to enhance both scalability and adaptability in dynamic web environments.

\section*{Ethical Considerations}
Branch-and-Browse aims to enhance autonomous web exploration in a controlled and responsible manner.  
All experiments were conducted in simulated environments (i.e., WebArena \cite{zhou2023webarena}) to avoid real-world data access or unintended interactions.  
However, when deployed on live websites, automated agents may raise concerns related to privacy, consent, and site policies.  
We recommend incorporating safeguards such as rate limiting, permission control, and transparent disclosure to ensure ethical and compliant use.
For this paper, large language models were used solely for text polishing and formatting assistance, not for idea generation or substantive content creation.

\section*{Acknowledgements}
We thank the reviewers, as well as members of SymbioticLab, for their helpful feedback.
This work was supported in part by NSF grants CCF-2450085 and CNS-2106184, and by grants from Ford and Cisco.



\bibliography{custom}

\clearpage
\appendix
\end{document}